\documentclass{article} % For LaTeX2e
\usepackage{iclr2019_conference,times}

\usepackage{amsmath,amsfonts,bm}

\def\eqref#1{equation~\ref{#1}}
\def\1{\bm{1}}

\DeclareMathAlphabet{\mathsfit}{\encodingdefault}{\sfdefault}{m}{sl}
\SetMathAlphabet{\mathsfit}{bold}{\encodingdefault}{\sfdefault}{bx}{n}

\newcommand{\softmax}{\mathrm{softmax}}

\usepackage{hyperref}
\usepackage{url}
\usepackage{latexsym}
\usepackage{graphicx}
\usepackage{amsmath}
\usepackage{arydshln}
\usepackage{url}
\usepackage{color}
\usepackage{algorithm}
\usepackage{todonotes}
\usepackage{enumitem}
\usepackage[noend]{algpseudocode}

\definecolor{mygreen}{rgb}{0,0.6,0}
\definecolor{mygray}{rgb}{0.5,0.5,0.5}
\definecolor{mymauve}{rgb}{0.58,0,0.82}

\newcommand{\SCRNN}{SC-RNN}
\newcommand{\SCHRNN}{SS-RNN}
\newcommand{\INDP}{INDP}
\newcommand{\crfRNN}{CRF}
\newcommand{\SimpleRNN}{RNN}
\newcommand{\OurRNN}{AC-RNN}
\newcommand{\SeqToSeq}{sequence-to-sequence}

\title{Efficient Sequence Labeling with Actor-Critic Training}
\author{Saeed Najafi$^{\dag}$ and Colin Cherry$^{\ddag}$ and Grzegorz Kondrak$^{\dag}$ \\
\begin{tabular}{cc}
 & \\ 
$^{\dag}$Department of Computing Science & $^{\ddag}$Google\\
University of Alberta, Canada     & Montreal, Canada\\
{\tt \{snajafi,gkondrak\}@ualberta.ca}    & {\tt colin.a.cherry@gmail.com}
\end{tabular}
}

\iclrfinalcopy
\begin{document}

\maketitle

\begin{abstract}
Neural approaches to sequence labeling often use a 
Conditional Random Field (CRF) to model their output
dependencies, while Recurrent Neural Networks (RNN) 
are used for the same purpose in other tasks.
We set out to establish RNNs as an attractive 
alternative to CRFs for sequence labeling.
To do so, we address one of the RNN's most prominent shortcomings, 
the fact that it is not exposed to its own errors with the maximum-likelihood training. 
We frame the prediction of the output sequence as a sequential decision-making process, %where the RNN takes a series of actions without being conditioned on the ground-truth labels.
where we train the network with an adjusted actor-critic algorithm (AC-RNN).
We comprehensively compare this strategy with 
maximum-likelihood training for both RNNs and CRFs on three structured-output tasks.
The proposed AC-RNN efficiently matches 
the performance of the CRF on NER and CCG tagging, and outperforms it on Machine Transliteration.
We also show that our training strategy 
is significantly better than other techniques for addressing RNN's exposure bias, 
such as Scheduled Sampling, 
and Self-Critical policy training.
\end{abstract}

\section{Introduction}
\label{intro}
Sequence labeling is a canonical structured output problem,
with many techniques available
to track its chain-structured output dependencies.
Conditional Random Fields (CRF)
built over a neural feature layer
have recently emerged as a best practice
for sequence labeling tasks in NLP~\citep{huang}.
Alternatively, one can track the same output dependencies
using a Recurrent Neural Network (RNN) over the output sequence,
similar to the decoder component in neural machine translation (NMT). Using a decoder RNN instead of a CRF has several potential advantages, 
such as simplified implementation,
tracking longer dependencies, and allowing for larger output vocabularies.
However, shifting from the CRF's sequence-level training objective
to the decoder RNN's sequence of token-level objectives may lead to 
suboptimal performance due to exposure bias.

Exposure bias stems from 
the token-level maximum likelihood objective
typically used in {\SimpleRNN} training, which
does not expose the model to its own errors~\citep{scheduled}.
RNNs are typically conditioned on gold-standard contexts during training 
(a procedure known as Teacher Forcing \citep{Goodfellow-et-al-2016}),
while at test time, the model is conditioned on its own predictions,
creating a train-test mismatch.
As sequence-level models, CRFs are immune to exposure bias.

We set out to establish the decoder RNN as an attractive alternative to the CRF for sequence labeling. We do so by adopting a simple and effective RNN training strategy to counter the exposure-bias problem, and by providing an experimental comparison to demonstrate that our modified RNN can match the CRF in accuracy, and surpass it in flexibility. Our chosen training method is Actor-Critic reinforcement learning~\citep{Konda2003}, which we adapt to the sequence-labeling scenario by providing immediate rewards after each output tag, and by letting those outputs adjust the critic scores. Our complete system, dubbed as the AC-RNN, maintains the same neural feature layer that has proven 
so successful for the CRF, changing only the output layer.\footnote{\url{https://github.com/SaeedNajafi/ac-tagger}} 
We conduct a comprehensive analysis comparing the {\OurRNN} to the {\crfRNN} under controlled
conditions using a shared implementation.
We demonstrate that on Named Entity Recognition (NER) and
Combinatory Categorical Grammar (CCG) supertagging,
the {\OurRNN} can match the performance of the {\crfRNN}, while training more efficiently.

To demonstrate its flexibility, we also test our method on machine transliteration,
a monotonic transduction problem that straddles the boundary between sequence 
labeling and full sequence-to-sequence modeling. Finally, we compare our method with previous 
techniques proposed to address exposure bias. 
On NER tagging, 
we empirically demonstrate that {\OurRNN} is 
significantly better than the {\SimpleRNN} trained with scheduled sampling \citep{scheduled}. 
We also demonstrate that our approach is more suitable for 
sequence labeling tasks than other similar policy-gradient methods 
such as self-critical training of \citet{Rennie2017SelfCriticalST}.

%The paper is organized as follows.
%Related works are summarized in Section \ref{prior_work}.
%Section~\ref{architecture} explains the base architecture of our models,
%providing the background information needed to introduce the adjusted Actor-Critic training in Section~\ref{novel}.
%Our comprehensive experiments are presented in Section~\ref{experiments}.

\section{Prior Work}
\label{prior_work}

%\begin{figure}
 %\centering
  %\includegraphics[width=8.0cm,height=1.80cm]{ccg-example}
 %\vspace*{-.2in} %GK: this lifts the caption to reduce white space 
 %\caption{CCG supertagging of the sentence \textit{`AirCanada Serves Edmonton'}.}
 %\label{ccg-example}
 %\end{figure}

In sequence labeling, several neural methods have recently been shown to 
outperform earlier systems that use hand-engineered features.
For the tasks of POS tagging, chunking and NER,
\citet{huang} apply a CRF output layer on top of a bi-directional RNN
over the source. For the NER task,
\citet{neuralarchitecture} extend the RNN layer with
character-level RNNs
that capture information about word prefixes and suffixes.
\citet{cnnNER} use Convolutional Neural Networks to a similar end.
We build upon Lample et al.'s approach, replacing their CRF with a 
decoder RNN trained with an adjusted Actor-Critic objective.

We also apply the {\OurRNN} to CCG supertagging \citep{C04-1041}, where the model
labels each word in a sentence with one of 1,284 morphosyntactic categories from CCGbank
\citep{Hockenmaier:2007:CCC:1288681.1288685}.
%an example is shown in Figure \ref{ccg-example}.
Bidirectional RNNs have been used in a number of recent supertagging
 systems \citep{N16-1026,D16-1181,N16-1027,Kadariarticle,AAAI1714723}. 
Among these, our effort is most similar to \citet{N16-1027}, 
who also use an RNN decoder over a
bidirectional encoder; however, they address exposure bias with Scheduled Sampling.

% \begin{figure}
  %\centering
  %\includegraphics[width=3.51cm,height=2.51cm]{tl-example}
  %\caption{Transliteration of the name \textit{`Osheen'} into Persian}
  %\label{tl-example}
  %\end{figure}
 
%GK! the following sentence is ungrammatical
We also consider transduction tasks, which go beyond sequence labeling
by allowing many-to-many monotonic alignments
between the source and target symbols.
In particular, we focus on transliteration,
where the goal is to convert a word from a source script to a target script
on the basis of the word's pronunciation. %(Figure \ref{tl-example}). 
Many neural transliteration approaches follow the {\SeqToSeq} model
originally proposed for NMT~\citep{DBLP:journals/corr/Jadidinejad16,DBLP:journals/corr/RoscaB16}.
Our {\OurRNN} transliteration system is similar to these,
but with an improved objective to address exposure bias.

Other approaches have been proposed to address exposure bias in {\SimpleRNN},
especially for NMT. We review the following major techniques: Scheduled Sampling, %Beam-search Training, %Professor Forcing,
Reinforcement Learning, and Imitation Learning. 
To our knowledge, none of these prior works has compared its method against CRF.

\noindent
\textbf{Scheduled Sampling (SS):} \citet{scheduled} introduce the notion of scheduled sampling as the decoder RNN is gradually exposed to its own errors, where
%In this approach, at each time step $t$ with probability $\epsilon$, 
%we feed the ground-truth token $y_{t-1}$ into the decoder RNN, 
%otherwise we use the model's greedily-generated token $\hat{y}_{t-1}$. 
a sampling probability is annealed at every training epoch so that 
we use gold-standard inputs at the beginning of the training, 
but while approaching the end, we instead condition the predictions 
on the model-generated inputs. 
%We consider this approach as one of our baselines. %Although Scheduled Sampling has been shown to outperform the Teacher Forcing maximum-likelihood objective on several sequence-to-sequence tasks \citep{scheduled}, it has been demonstrated that as $\epsilon$ approaches zero, the model learns a adjusted conditional distribution which is different than the true distribution \citep{huszar2015not}.

%\noindent
%\textbf{Beam-search Training:} \citet{rush} employ beam search in the training phase where the gold sequence remains at the top of the beam list at each step. In this approach, if the gold sequence is not ranked first, the method recognizes a violation, and its corresponding cost is back-propagated to the model's parameters. This method linearly increases the training time with respect to the beam size.

%\noindent
%\textbf{Professor Forcing:} \citet{NIPS2016_6099} introduce the Professor Forcing method based on the Generative Adversarial Networks \citep{Goodfellow:2014:GAN:2969033.2969125} to encourage the dynamics (parameters hidden space) of the decoder RNN to be the same during training and testing (the discriminator is fooled accordingly), however, they report lower results compared to the Teacher Forcing technique on shorter sequences with length less than 100.

\noindent
\textbf{Reinforcement Learning (RL):} %Reinforcement learning has been applied 
%to structured prediction tasks \citep{Maes:2009:SPR:1666220.1666240}. 
%One of our contributions is to take the 
%formalism of Structured Prediction Markov-Decision Process (SP-MDP) 
%introduced by \citet{Maes:2009:SPR:1666220.1666240}, 
%and apply it to deep neural architectures, whereas \citet{Maes:2009:SPR:1666220.1666240} 
%only investigated non-neural feature-based methods.
\citet{mixer} apply the REINFORCE algorithm \citep{reinforce} to 
Neural Machine Translation (NMT), to train the network with a 
reward derived from the BLEU score of each generated sequence. 
\citet{actor-critic} apply the actor-critic algorithm in NMT by applying a 
reward-reshaping approach to construct intermediate BLEU feedback
 at each step. \citet{Rennie2017SelfCriticalST} introduce a Self-Critical (SC) 
 training approach that does not require a critic model, which has been 
 shown to outperform REINFORCE in the image captioning task. 
 This method has also been applied to abstract summarization \citep{DBLP:journals/corr/PaulusXS17}. 

%The SC training is intended to represent the state-of-the-art in 
%reinforcement learning for sequence-to-sequence models with 
%sequence-level rewards, which will be another baseline in our experiments.
%The SC training samples a token based on the probability distribution 
%formed over all tokens at each step, resulting in the sampled sequence 
%$\tilde{Y}=(\tilde{y}_{1}, ..., \tilde{y}_{l'})$. The $\tilde{Y}$ is 
%then compared to the greedily-sampled sequence $\hat{Y}=(\hat{y}_{1}, ..., \hat{y}_{l'})$. 
%The SC defines a sequence-level cost between the sampled sequences 
%and the gold prediction $Y$.
%If the sampled sequence $\tilde{Y}$ has a lower cost than the greedily-sampled sequence, 
%the likelihood of $\tilde{Y}$ will be increased.

Unlike these previous works that apply reinforcement-learning techniques to 
optimize an available external metric such as ROUGE in text summarization, 
or BLEU in translation, giving one reward at the end of each sequence, we 
demonstrate that sequence labeling tasks benefit from the binary rewards 
that are available at each step. In addition, \citet{actor-critic}, and \citet{DBLP:journals/corr/PaulusXS17} 
combine the Teacher Forcing maximum-likelihood objective with their 
proposed RL objectives, which requires two forward computations in the decoder RNN, 
one for conditioning on the ground-truth labels, another for the RL objective without 
conditioning on the ground-truth labels.
In this work, we will incorporate the supervision of the gold label into the actor-critic 
algorithm itself without any extra computation. In contrast to \citet{actor-critic}, our method 
employs a simpler critic architecture, without any schedules to pre-train the critic model. %where \citet{actor-critic} uses separate encoder-decoder 
%RNNs for the critic model, which doubles the parameters of the network. 
%Moreover, our approach will not rely on any schedules to pre-train the critic model. 
%We will also present the first direct, controlled comparison between CRFs and any form of RNN.

\noindent
\textbf{Imitation Learning:}
%The Scheduled Sampling is a special variant of the search-based methods 
%designed for structured prediction problems \citep{daume06thesis,daume09searn, ross2011reduction,changb15}. 
%The Scheduled Sampling can also be considered as a special variant of this meta-learning approach. 
%The main idea behind it is to transform the structured prediction into a simpler multi-class classification problem. To
%achieve this, \citet{daume09searn} propose in their SEARN algorithm to train a local classifier
%to predict each token sequentially, thus searching step by step in the big 
%combinatorial space of the structured outputs. 

Another alternative method to address exposure bias would be imitation learning, where one has access to a gold-standard policy instead of rewards. In the case of sequence labeling, this policy corresponds to the gold-standard tag sequence, and most imitation-learning algorithms such as Dagger \citep{Ross2011ARO} reduce to variants of Scheduled Sampling. Recently, the related technique of learning to search \citep{daume09searn} has been extended to neural network models by SEARNN \citep{leblond2018searnn}, but that has been shown to perform slightly worse than the actor-critic training for NMT.
\section{Architecture}
\label{architecture}

%\begin{figure}[t]
 %\centering
  %\includegraphics[width=8.0cm,height=4.52cm]{tagging-encoder}
% \vspace*{-.2in} %GK: this lifts the caption to reduce white space
% \caption{Time-unfolded encoder for labeling the sentence \textit{`I went to the Grand Mall'}. $F$ retrieves the computed feature vectors. The symbol `+' denotes the concatenation of the forward and backward outputs.}
 %\label{tagger-encoder}
 %\end{figure}
 
%We are motivated by the observation that both CRFs and RNNs
%can be used to track the output dependencies needed to generate coherent sequences.
%In an analogy to encoder-decoder models,
%either could be used as a decoder.
%Following that analogy,
%we also need an encoder to build input features,
%which we share across all competing systems.

\subsection{Encoder}
\label{encoder}

Following \citet{huang},
the encoder of our sequence labeler employs
a bi-directional RNN over the tokens in the sequence.
The bi-directional RNN transforms context-independent token representations
into representations of tokens-in-context,
allowing each position to potentially encode information
from the entire input sequence.

For sequence labeling,
we follow \citet{neuralarchitecture}, and
build our context-independent word representation by combining
an embedding table
with the outputs of a bi-directional RNN applied to each word's characters. % (Figures \ref{tagger-encoder} and \ref{prefix-suffix-encoder}).
The final states of the forward and backward character RNNs are
concatenated to the word's embedding, and then passed through a dropout layer.
%GK! what is the reason for deviating?
%Deviating slightly from previous work, 
We also concatenate capitalization pattern indicators
to these feature vectors.
Our word embeddings are initialized using embeddings pre-trained on a large corpus.

For transliteration, where we operate exclusively on the character level, we apply a bi-directional RNN on
the character representations, which are provided by a randomly initialized embedding table.

%\begin{figure}[t]
 %\centering
  %\includegraphics[width=7.0cm,height=5.5cm]{prefix-suffix-encoder}
 %\caption{Encoder for building the feature vector of `Mall'. $E$ retrieves word or character embeddings.}
% \label{prefix-suffix-encoder}
 %\end{figure}
 
%\begin{figure}
% \centering
%  \includegraphics[width=8.5cm,height=4.86cm]{tl-encoder}
% \caption{Time-unfolded encoder for transliterating \textit{`Osheen'}.}
% \label{tl-encoder}
% \end{figure}

\subsection{Decoders}
\label{decoder}
Given an input $X=(x_{1}, x_{2}, ..., x_{l})$,
we look for an output sequence $Y=(y_{1}, y_{2}, ..., y_{l})$ where each
$y_{t}$ is an output token.
In the encoder (Section~\ref{encoder}),
we transform the input $X$ into a sequence of hidden vectors $H=(h_{1}, h_{2}, ..., h_{l})$.
Given these vectors,
the simplest decoder does not account for output dependencies at all.
Instead, it independently
predicts the output at time $t$ by mapping $h_{t}$
into a probability distribution $p_{\mathrm{INDP}}(y_t | h_t)$ using a $\softmax$ layer.
This results in a sequence-level probability of
%GK! p_INDP is undefined
$p(Y|X)=\prod_{t} p_{\mathrm{INDP}}(y_t | h_t)$.
%where $p_{\mathrm{INDP}}(y_t | h_t) = \frac{e^{S_{t,y_{t}}}}{\sum_{y_{j}} e^{S_{t,y_{j}}}}$ and $S_{t,y_{j}}$ is the score of tag $y_{j}$ at time $t$.

In sequence labeling tasks,
there are typically dependencies between the output tokens;
for example,
in English Part of Speech tagging, a determiner is unlikely to follow another determiner.
The most prominent and widely used approach to track such dependencies is the CRF,
which uses dynamic programming over an undirected graphical model
to maintain a well-defined probability distribution over sequences,
effectively modeling
$p(Y|X)=p_{\mathrm{CRF}}(Y|H)$,
where $p_{\mathrm{CRF}}$ hides fixed-order Markov dependencies over $Y$.
The CRF can be used as a node in a neural sequence labeler,
while still allowing the system to train end-to-end~\citep{huang}.

An alternative technique is to use a decoder RNN on top of the encoder,
as is typically done in neural machine translation~\citep{NIPS2014_5346},
and has been done for CCG super-tagging~\citep{N16-1027}.
% CC: Easy, math-light verison here
% The decoder RNN generates a state $d_t$ from the previous output token $y_{t-1}$ and the previous hidden state $d_{t-1}$.
% This state is then concatenated with a context vector $c_{t}$
% that summarizes the input $X$ for the current time step.
% For sequence labeling, the source-to-target alignment is trivial, and
% $c_t$ is provided by the encoder representation at time $t$: $c_t=h_t$.
% Finally, a $\softmax$ layer is used to define a probability distribution over the
% output tokens, resulting in a model of:
% $p(Y|X)=\prod_{t} p_{\mathrm{RNN}}(y_t|c_t, y_{t'<t})$.
% During training, the gold-standard previous token $y_{t-1}$ is fed into the decoder at time $t$,
% while at test time, we use the model's output.
%
%
% CC: Math-heavy version
Let $d_t$ be the recurrent decoder state, summarizing the output sequence up to time $t$,
and let $c_t$ be the context vector that summarizes the input $X$ for time $t$.
For sequence labeling, the source-to-target alignment is trivial, and
$c_t$ is provided directly by the encoder: $c_t=h_t$.
We can then use the input-feeding method of \citet{attention} to define $d_t$ recursively:
\mbox{$d_{t} = \mathrm{RNN}(d_{t-1}, y_{t-1}, c_{t-1})$}, where $\mathrm{RNN}$ is a recurrent unit, in our case, an LSTM~\citep{Hochreiter1997}.
Finally, a $\softmax$ layer is used to define a probability distribution over
output tokens $p_{\mathrm{SM}}(y_t|c_t, d_t)$,
resulting in a sequence model of:
$p(Y|X)=\prod_{t} p_{\mathrm{RNN}}(y_t|h_t, y_{t'<t}) = \prod_t p_{\mathrm{SM}}(y_t|c_t, d_t)$.
During training, the gold-standard previous token $y_{t-1}$ is fed into the decoder at time $t$,
while at test time, we use the model's generated output.

For transliteration, where the input and output sequence lengths do not match,
we can no longer simply provide the encoder state $h_t$ at time $t$ as the context vector $c_{t}$ for our probability models.
Following standard practice in NMT, for the decoder RNN,
an attention mechanism~\citep{atten-bah} can learn an alignment model that
provides a scalar score $\alpha_{t,t'}$ for each target position $t$ and source position $t'$, giving us a context vector \mbox{$c_t = \sum_{t'} \alpha_{t,t'} h_{t'}$},
which we can use in place of $h_t$ in the RNN models described above.
Specifically, we use the global-general attention mechanism of \citet{attention}.
%For the CRF, an output hidden state is not available as conditioning information for learned attention.
%Instead, we build context vectors $c_{t} =(h_{t-1}, h_{t}, h_{t+1})$.
%These concatenated vectors act as
%a window-based local attention~\citep{attention}.
To modify the CRF for transliteration, we must allow it to generate output of a different length
from its input. To do so, we pad both sequences with extra end symbols up to a fixed maximum length,
and let CRF decode until the end of the padded source sequence.
It controls its target length by outputting padding tokens.

All of the models described above can be trained with a maximum likelihood objective:
\begin{center}
$J_{ml}(\theta)=\sum_{X,Y} \ln p_{\theta}(Y|X)$
\end{center}

\section{Actor-Critic Training}
\label{novel}

We adopt the actor-critic algorithm \citep{RLbook, Konda2003,DBLP:journals/corr/MnihBMGLHSK16} to fine-tune the decoder RNN.
In AC training,
the decoder RNN first generates a greedy output sequence
according to its current model,
similar to how it would during testing.
We calculate a sequence-level credit (return) for each prediction by comparing it to the gold-standard.
The AC update modifies our RNN to improve credits at each step.
This process exposes the decoder to its own errors,
alleviating exposure bias.
Algorithm \ref{alg:acupdate} provides pseudo code for the training process,
which we expand upon in the following paragraphs.

We define the token-level reward $r_{t}$ as $+1$
if the generated token $\hat{y}_{t}$ is the same as the gold token $y_{t}$,
and as $0$ otherwise.
We compute the sequence-level credit $G_{t}$ for each decoding step
using the multi-step Temporal Difference return~\citep{RLbook}:
\begin{center}
$G_{t} = \sum_{i=0}^{n-1} \left[r_{t+i}\right] + V_{\theta^{'}}(t+n)$
\end{center}

\begin{algorithm}[t]
\caption{adjusted Actor-Critic Training}
\label{alg:acupdate}
\begin{itemize}[label=$\cdot$] \itemsep -0.1cm
\item Input: Source $X$, Target $Y$, and $n$ as hyper-parameter
\item Greedy decode $X$ using $\theta$ to get:
  \begin{itemize}[label=$\cdot$]
  \item the output sequence $\hat{Y}=(\hat{y}_{1}, ..., \hat{y}_{l})$
  \item decoder RNN states $D = (d_{1}, ..., d_{l})$
  \item context vectors $C = (c_{1}, ..., c_{l})$
  \end{itemize}
\item For each output target position $t$:
  \begin{itemize}[label=$\cdot$]
  \item $r_t=1$ if $\hat{y}_t=y_t$, 0 otherwise
  \item $V_{\theta'}(t) = \mathrm{CriticNetwork}(d_t,c_t, \theta')$
  \end{itemize}
\item $\mathit{loss}_{\theta}=0$; $\mathit{loss}_{\theta'}=0$
\item For each output target position $t$:
  \begin{itemize}[label=$\cdot$]
  \item $G_{t} = \sum_{i=0}^{n-1} \left[r_{t+i}\right] + V_{\theta^{'}}(t+n)$
  \item $\delta_t =  G_{t} - V_{\theta'}(t)$
  \item $a\delta_t = \textrm{adjust}( y_t, \hat{y}_t, \delta_t) \times \delta_t$
  \item $\mathit{loss}_{\theta} = \mathit{loss}_{\theta} - a\delta_t \ln p_\theta(\hat{y}_t|X,\hat{y}_{t'<t})$
  \item $\mathit{loss}_{\theta'} = \mathit{loss}_{\theta'} + \delta_t \times \delta_t$
  \end{itemize}
\item Back-propagate through $\mathit{loss}_{\theta}$ as normal to update $\theta$
\item Perform a semi-gradient step along loss $\theta'$ to update $\theta'$
\end{itemize}
\end{algorithm}

%
%\begin{table*}[ht]
  %\centering
  %\small
%\begin{tabular}{ccccc}
 %Input phrase & the & University & of & XYZ\\
 %Reference tags & O & Org & Org & Org\\
 %Model's prediction & O & Org & O & Loc\\
%$J_{ml}$ & \multicolumn{1}{r}{$\ln P(\Out)$} & \multicolumn{1}{r}{$\ln P(\Org | \Out)$} & \multicolumn{1}{r}{$\ln P(\Org | \Out \; \Org)$} & \multicolumn{1}{r}{$\ln P(\Org | \Out \; \Org \; \Org)$}\\
%$J_{ac}$ & $1.9\ln P(\Out)$ & $0.9\ln P(\Org | \Out)$ & $-0.1\ln P(\Out | \Out \; \Org)$ & $-0.1\ln P(\Loc | \Out \; \Org \; \Out)$\\
%\end{tabular}
%\caption{The formed probabilities by maximum-likelihood ($J_{ml}$) and actor-critic ($J_{ac}$) objectives for named entity tagging of the phrase `the University of XYZ'. The advantage terms $\delta_t=1.9$, $0.9$, $\ldots$ given in $J_{ac}$ are computed using $n=4$, and $V(t)=0.1$.}
%\label{ac-example}
%\end{table*}

The step count $n$ allows us to control our bias-variance trade-off,
with a large $n$ resulting in less bias but higher variance. 
The critic $V_{\theta^{'}}(t)$ is a regression model that estimates the expected return $E\left[G_{t}\right]$, taking the context vector $c_{t}$ and the decoder's hidden state $d_{t}$ as input. %\footnote{We assume the current state of the environment is observed by the vectors $c_{t}$, which summarizes the relevant input tokens, and the decoder state $d_{t}$, which summarizes the previously generated output tokens.}
It is trained jointly alongside our decoder RNN, using a distinct optimizer (without back-propagating errors through $c_{t}$ and $d_{t}$).
With this critic in place, the update for the AC algorithm
is defined as
\begin{center}
$\frac{\partial J_{ac}(\theta)}{\partial \theta} = \sum_{t} \frac{\partial \log(p_{\theta}(\hat{y}_{t}|X, \hat{y}_{t'<t}))}{\partial \theta} \left(\delta_{t}\right)$
\end{center}
where:
\[
\delta_t = G_{t} - V_{\theta'}(t)
\]
The AC update changes the prediction likelihood proportionally to the advantage $\delta_{t}$ of the token $\hat{y}_{t}$.
Therefore, if $G_{t} > V_{\theta^{'}}(t)$, the decoder should increase the likelihood.
The AC error $\frac{\partial J_{ac}(\theta)}{\partial \theta}$ back-propagates only through the actor's prediction likelihood $p_{\theta}$.

%Table \ref{ac-example} illustrates an example in NER, where the model tends to incorrectly label an entity as `Location' instead of `Organization'. We directly give negative credits for the invalid predictions with the $J_{ac}$ objective.

\noindent
\textbf{Critic Architecture}:
We employ a non-linear feed-forward neural network as our critic, which uses leaky-ReLU activation functions \citep{Nair:2010:RLU:3104322.3104425} in the first two hidden layers. In the output layer, it uses a linear transformation to generate a scalar value. To learn the critic's parameters $\theta'$, we use a semi-gradient update \citep{RLbook}. We do not use the full gradient in the Mean Squared error to train this regression model. Accordingly, for $\frac{\partial \mathit{loss}_{\theta'}}{\partial \theta'} = \frac{\partial (\delta_t \times \delta_t)}{\partial \theta'}$, 
instead of using $2 \delta_t \frac{\partial (G_{t} - V_{\theta'}(t))}{\partial \theta'}$, we use the update $2 \delta_t \frac{\partial V_{\theta'}(t)}{\partial \theta'}$. The Temporal Difference return $G_{t}$ uses the critic's estimate $V_{\theta^{'}}(t+n)$. The full gradient will cause a feedback loop as by doing so, we will allow $G_{t}$ to match with $V_{\theta'}(t)$ in order to reduce the Mean Squared error.

\noindent
\textbf{Adjusted Training}:
Due to the inevitable regression error of the critic, and the fact that it is randomly initialized at the beginning,
the advantage $\delta_t$ can undesirably become negative for a correctly-selected tag, or positive for a wrongly-selected tag. Optimizing the network according to these invalid advantages would increase the probability of the wrong tags, while decreasing the probability of the true tag. 
In such cases, to help critic update itself and form better estimates in the next iteration, we clip $\delta_t$ to zero by defining the adjusted advantage $a\delta_t$ as $\mathrm{adjust}( y_t, \hat{y}_t, \delta_t)$ $\times$ $\delta_t$ where:
\begin{center}
$\mathrm{adjust}( y_t, \hat{y}_t, \delta_t) = 
\begin{cases}
0 \quad \mbox{ if } \quad \hat{y}_{t} = y_{t} \quad \& \quad \delta_t < 0\\
0 \quad \mbox{ if } \quad \hat{y}_{t} \neq y_{t} \quad \& \quad \delta_t > 0\\
1 \quad otherwise \\
\end{cases}$
\end{center}
By setting the advantage $a\delta_t$ to 0, the $\mathrm{adjust}$ term effectively switches off the entire actor update
when the advantage has the wrong polarity.
%This introduces bias, but also serves to stabilize training.
Note that the critic is always updated. 
%The adjusted training can also be interpreted as a combination of two advantages $\delta_t + \delta'_t$, where:
%\begin{center}
%$\delta'_t =
%\begin{cases}
 %-\delta_t \quad \mbox{ if } \quad \hat{y}_{t} = y_{t} \quad \& \quad \delta_t < 0\\
 %-\delta_t \quad \mbox{ if } \quad \hat{y}_{t} \neq y_{t} \quad \& \quad \delta_t > 0\\
%0 \quad otherwise 
%\end{cases}$
%\end{center}
%The $\delta'_t$ adds the supervision of the gold label into the objective as it is always positive for the correctly selected tag, and negative for the wrongly selected token.

\begin{figure}
 \centering
  \includegraphics[width=10.00cm,height=6.5cm]{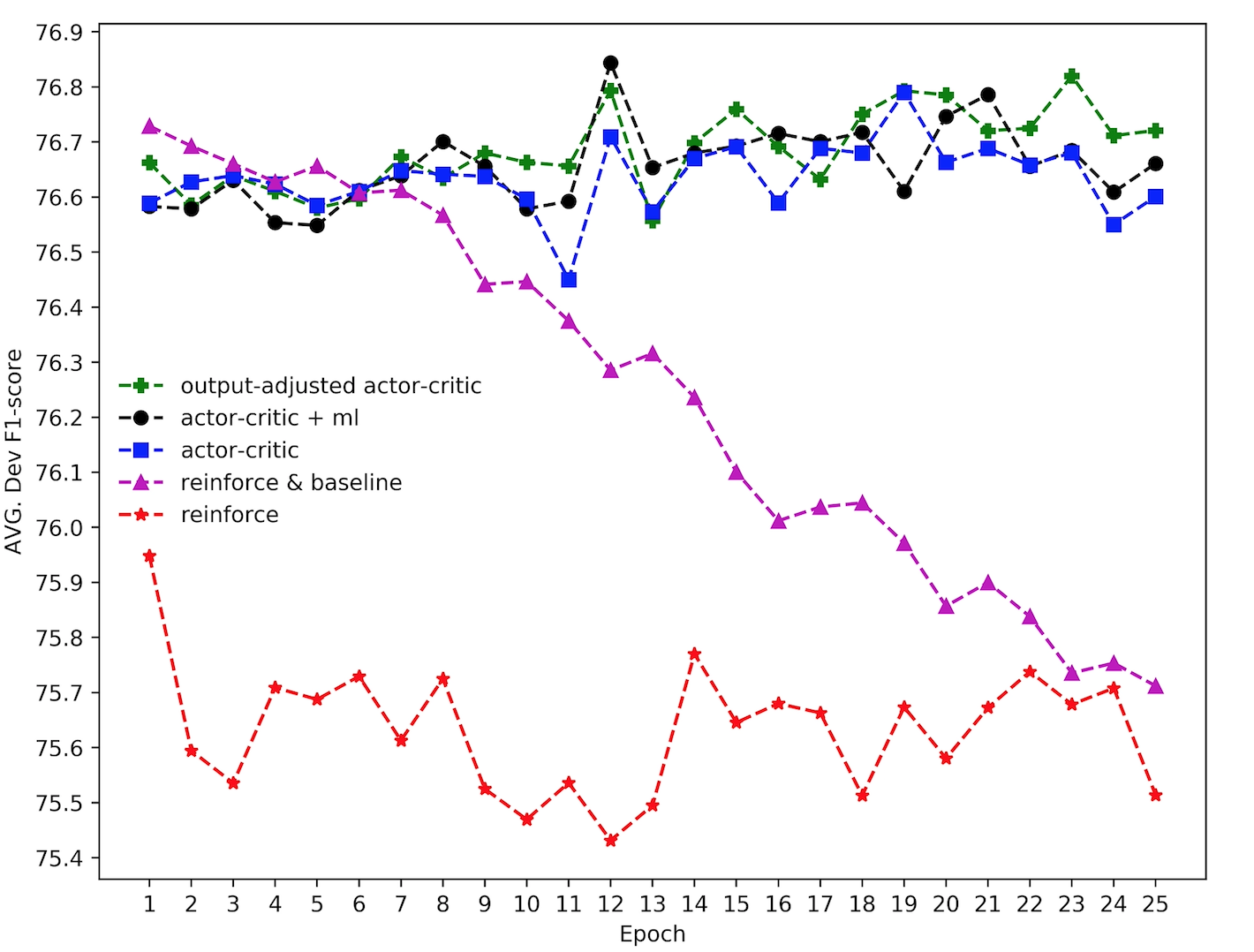}
   \vspace*{-.2in} %GK: this lifts the caption to reduce white space
 \caption[The adjusted actor-critic objective compared to other methods]{The adjusted actor-critic objective compared to alternative policy-gradient objectives on German NER with 17 possible output tags. All objectives are maximized using Gradient Ascent with a fixed step size of 0.5 for actor-critic, and 0.01 for REINFORCE objectives. For REINFORCE objectives, we set $n=l$ in the Temporal Difference credits (i.e. sum all rewards until the end of sequence). All methods are trained 20 times with different random seeds using the same hyper-parameters.}
 \label{ac-bac-comparison}
 \end{figure}
 
Similar to prior RL works \citep{mixer, actor-critic, DBLP:journals/corr/PaulusXS17}, we pre-train the network using the Teacher Forcing maximum-likelihood objective ($J_{ml}$). 
We then continue training from the best model
using our adjusted actor-critic objective. Figure \ref{ac-bac-comparison} shows development set performance, starting from the same $J_{ml}$-pre-trained point, for our adjusted objective, 
as compared to standard actor-critic, REINFORCE with baseline, and normal REINFORCE on German NER. We observe that the adjusted training helps the model reach a higher point compared to all other objectives.  Unlike prior RL methods, the adjusted actor-critic objective does not require any schedule for pre-training the critic. It also avoids the necessity of combining the actor-critic objective with the Teacher Forcing maximum-likelihood training, which is shown as actor-critic + ml in Figure \ref{ac-bac-comparison}. After the pre-training phase, the related works \citep{mixer, actor-critic, DBLP:journals/corr/PaulusXS17} still combine the maximum-likelihood training with their proposed RL objectives.
 
\section{Experiments}
\label{experiments}
We comprehensively compare {\OurRNN}, {\crfRNN}, and {\SimpleRNN} on
three tasks: NER tagging, CCG supertagging, and machine Transliteration. We then compare the adjusted actor-critic objective with Scheduled Sampling on NER. Finally, we compare our training strategy to the Self-Critical method of \citet{Rennie2017SelfCriticalST}.
\subsection{Setup}

\noindent
\textbf{Datasets}:
To conduct the NER experiments, we use the English and German datasets of the CoNLL-2003 shared task \citep{2003conll}. Both datasets are annotated with 4 different entity types: `Location', `Organization', `Person', and `Miscellaneous' (e.g. events, nationalities, etc.). As we have multi-word named entities (e.g. `University of XYZ'), we employ the `BILOU' tagging scheme \citep{overview2009}. %\footnote{Also known as the `IOBES' tagging scheme.} \citep{overview2009}.

For CCG supertagging, we use the English CCGbank \citep{Hockenmaier:2007:CCC:1288681.1288685}, the standard sections \{02-21\},  \{00\}, and \{23\}
as the train, development, and test sets, respectively.
We consider all the 1284 supertags appeared in the train set.

We use pre-trained, 100-dimensional \textit{Glove} embeddings \citep{pennington2014glove} for all English word-level tasks,
and fine-tune them during training. For German NER, we obtain the embeddings (64 dimensions) of \citet{neuralarchitecture}, which are trained on a German monolingual dataset from the 2010 Machine
Translation Workshop.
We apply no preprocessing on the datasets except replacing
the numbers and unknown words with the `NUM' and `UNK' symbols.

We conduct the transliteration experiments on
the English-to-Chinese (EnCh), English-to-Japanese (EnJa), English-to-Persian (EnPe), and English-to-Thai (EnTh) datasets of the NEWS-2018 shared task.\footnote{http://workshop.colips.org/news2018/shared.html}
The training sets contain approximately 40K, 30K, 10K and 30K instances
for EnCh, EnJa, EnPe, and EnTh, respectively, while the development
sets have 1K instances.
We train the models on the training sets,
and evaluate them on the development sets.
We hold out 10\% of the training sets as our internal tuning sets.

%
%\begin{table}
%\centering
%\small
%\begin{tabular}{ |c|c|c|c|c|c| }
 %\hline
% Split & Sentences & LOC & ORG & PER & MISC\\
 %\hline
 %Training set & 14987 & 7140 & 6321 & 6600 & 3438 \\
% Dev set & 3466 & 1837 & 1341 & 1842 & 922 \\
 %Test set & 3684 & 1668 & 1661 & 1617 & 702 \\
%\hline
%\end{tabular}
%\caption[The number of sentences and named entities for English]{The number of sentences and annotated named entities on the CoNLL-2003 English dataset.}
%\label{en-ner-data-stat}
%\end{table}

%\begin{table}
%\centering
%\small
%\begin{tabular}{ |c|c|c|c|c|c| }
 %\hline
% Split & Sentences & LOC & ORG & PER & MISC\\
 %\hline
 %Training set & 12705 & 4363 & 2427 & 2773 & 2288 \\
% Dev set & 3068 & 1181 & 1241 & 1401 & 1010 \\
 %Test set & 3160 & 1035 & 773 & 1195 & 670 \\
%\hline
%\end{tabular}
%\caption[The number of sentences and named entities for German]{The number of sentences and annotated named entities on the CoNLL-2003 German dataset.}
%\label{de-ner-data-stat}
%\end{table}

\noindent
\textbf{Training Details}: 
For the experiments, our different models share the same encoder, using the same number of hidden units (see the appendix for the hyper-parameters used in our experiments).
The maximum-likelihood training is done with the Adam optimizer \citep{DBLP:journals/corr/KingmaB14} with a learning rate of 0.0005. The RL training is done with the mini-batch gradient ascent ($\theta = \theta + \alpha \frac{\partial J_{ac}(\theta)}{\partial \theta}$) using a fixed step size of 0.5 for NER \& CCG, and 0.1 for Transliteration \footnote{We also tried Adam and RMSProp optimizers for the RL training, but both completely diverged.}. The critic is trained with a separate Adam optimizer with the learning rate of 0.0005. We employ a linear-chain first-order undirected graph in the CRF model.

As performance varies depending on the random initialization,
we train each model 20 times for NER and 5 times for CCG using different random seeds which are the same for all models. 
We report scores averaged across these runs $\pm$ the standard deviations.
Due to time constraints, for the transliteration experiments, we train each model only once.\\

\noindent
\textbf{Evaluation}:
We compute the standard evaluation metric for each task:
entity-level F1-score for NER, tagging accuracy for CCG, and word-level accuracy for transliteration. %\footnote{A multi-word entity is predicted correctly if all of its words have the correct labels.} 
For the models with decoder RNNs, we report the results achieved using a beam search with a beam of size 10.
For the NER and CCG experiments,
we conduct the significance tests on the unseen final test sets, using the Student's t-test over random replications 
at the significance level of $0.05$.

\subsection{Results}

\noindent
\textbf{Main Comparisons}:
As our primary empirical study, we compare the {\OurRNN} to {\crfRNN} and {\SimpleRNN} models.
we also consider independent prediction of the labels 
as another baseline.

The results of the NER experiments are shown in Tables \ref{ner}.
As expected, we observe that by modelling the output dependencies using either an RNN or a CRF, we achieve a significant improvement over the baseline {\INDP}, about $1\%$ F1-score on both English and German datasets. 
With respect to prior work, our CRF model replicates the reported results on English NER. On German NER, we cannot replicate the CRF results of \citet{neuralarchitecture}, although we obtained their German word embeddings. We attribute this discrepancy to different preprocessing of the dataset.
Moreover, {\OurRNN} significantly outperforms both {\SimpleRNN} and {\crfRNN} on both English and German test sets with the corresponding P values of 0.001 and 0.004 for {\SimpleRNN}, and 0.003 and 0.016 for {\crfRNN}.
These results demonstrate that 
{\OurRNN} is successful at  overcoming the RNN's exposure bias, 
and represents a strong alternative to {\crfRNN} 
for named entity recognition.

\begin{table}
\centering
\begin{tabular}{ c c c | c c}
\hline
Model & Dev (En) & Test (En) & Dev (De) & Test (De) \\
 \hline
{\INDP} & 93.63 \small{$\pm$0.13} & 89.77 \small{$\pm$0.21} & 75.51 \small{$\pm$0.28} & 72.15 \small{$\pm$0.57} \\ 
{\SimpleRNN} & 94.43 \small{$\pm$0.16} & 90.75 \small{$\pm$0.23} & 76.85 \small{$\pm$0.39} & 73.52 \small{$\pm$0.36}\\
{\crfRNN} & 94.47 \small{$\pm$0.12} & 90.80 \small{$\pm$0.19} & 76.27 \small{$\pm$0.35} & 73.59 \small{$\pm$0.36}\\
{\OurRNN} & \textbf{94.54} \small{$\pm$0.12} & \textbf{90.96} \small{$\pm$0.15} & \textbf{77.10} \small{$\pm$0.29} & \textbf{73.82} \small{$\pm$0.29}\\
\hline
\citet{neuralarchitecture} & & 90.94 &  & 78.76\\
\hline
\end{tabular}
\caption{Average entity-level F1-score for English \& German NER on the CoNLL-2003 datasets. 
\citet{reimers-gurevych:2017:EMNLP2017} report 90.81 as the median performance for the CRF model of \citet{neuralarchitecture} in English.} 
\label{ner}
\end{table}

%\begin{table}
%\centering
%\begin{tabular}{ c c c}
%\hline
%Model & Dev & Test \\
 %\hline
%{\INDP} & 75.51 \small{$\pm$0.28} & 72.15 \small{$\pm$0.57} \\ 
%{\SimpleRNN} & 76.85 \small{$\pm$0.39} & 73.52 \small{$\pm$0.36} \\
%{\crfRNN} & 76.27 \small{$\pm$0.35} & 73.59 \small{$\pm$0.36} \\
%{\OurRNN} & \textbf{77.10} \small{$\pm$0.29} & \textbf{73.82} \small{$\pm$0.29} \\
%\hline
%\citet{neuralarchitecture} & & 78.76 \\
%\hline
%\end{tabular}
%\caption[The German NER results]{Average entity-level F1-score for German NER on the CoNLL-2003 dataset.} 
%\label{de-ner}
%\end{table}

On CCG supertagging (Table \ref{ccg}), 
{\OurRNN} is significantly better than all other models with the P values of 0.019, 0.025, and 0.002, respectively, and is competitive with reported state-of-the-art results.
For this task, we had expected the improvements to be larger,
because of CCG supertagging's potential for long-distance output dependencies.
Instead, the results show that independent predictions do surprisingly well.

%The CCG experiments reveal that {\OurRNN} is trained more efficiently than CRF.
Table~\ref{ccg} also shows the time and memory requirements for each method on the CCG task.
We observe that,
due to the large output vocabulary size of the task (1284 supertags),
CRF is five times slower than {\OurRNN} during training, 
while the batched version of its Forward algorithm requires six times more GPU memory during training. The Forward algorithm of CRF runs out of memory with the mini-batch size of 16 on a 12-GB Graphical Processing Unit.

\begin{table}
\centering
\begin{tabular}{ c c c | c c}
\hline
 Model & Dev & Test & Memory (GB) & Time (m)\\
 \hline
{\INDP} & 94.24 \small{$\pm$0.03} & 94.25 \small{$\pm$0.11} & \textbf{1.3} & \textbf{5} \\
{\SimpleRNN} & 94.25 \small{$\pm$0.06} & 94.28 \small{$\pm$0.09} & 1.8 & 11 \\
{\crfRNN} & 94.31 \small{$\pm$0.07} & 94.15 \small{$\pm$0.11} & 9.8 & 50 \\
{\OurRNN} & \textbf{94.43} \small{$\pm$0.08} & \textbf{94.39} \small{$\pm$0.06} & 1.5 & 10 \\
\hline
\citet{N16-1027} & 94.24 & 94.50 & & \\
\citet{Kadariarticle} & 94.37$^\diamond$ & 94.49$^\diamond$ & &\\
\hline
\end{tabular}
\caption[The CCG supertagging results]{Average top-1 accuracy, and the required GPU memory and execution time (one epoch) on English CCG supertagging. The $^\diamond$ results are achieved with CRF training.}
\label{ccg}
\end{table}

%\begin{table}
%\centering
%\begin{tabular}{ c c c}
%\hline
% Model & Memory (GB) & Time (m)\\
 %\hline
%{\INDP} & \textbf{1.3} & \textbf{5}\\
%{\SimpleRNN} & 1.8 & 11 \\
%{\crfRNN} & 9.8 & 50 \\
%{\OurRNN} & 1.5 & 10 \\
%\hline
%\end{tabular}
%\caption[The efficiency of AC-RNN compared to CRF]{The required GPU memory and execution time for one training epoch of the models on CCG supertagging using mini-batches of size 10.}
%\label{time-ccg}
%\end{table}

The transliteration results in Table \ref{dev-trans}
show that
{\OurRNN} outperforms {\crfRNN}
(likely due to {\crfRNN}'s inability to predict an output of a different length from its input),
as well as {\SimpleRNN} (likely due to its exposure bias).
The transliteration experiments support our hypothesis that 
{\OurRNN} is more generally-applicable than {\crfRNN},
and the improvements from the adjusted actor-critic training transfer to other tasks.

\begin{table}
\centering
\begin{tabular}{ c c c c c }
\hline
Model & EnCh & EnJa & EnPe & EnTh\\
\hline
{\crfRNN} & 67.6 & 45.8 & 75.6 & 32.2 \\
{\SimpleRNN} & 70.6  & 51.6 & 76.3 & 39.7\\
{\SCRNN} & 70.2  & 51.8 & 77.2 & 41.3\\
{\OurRNN} & \textbf{72.3} & \textbf{52.4} & \textbf{77.8} & \textbf{41.4} \\
\hline
OpenNMT & 70.1 & 47.7 & 70.5 & 36.3 \\
\hline
\end{tabular}
\caption[The transliteration results]{The word-level transliteration accuracy on the development sets of NEWS-2018 shared task. SC: Self-Critical training}
\label{dev-trans}
\end{table}

To confirm that our RNN baseline performs reasonably well, 
we validate our transliteration model against a standard NMT implementation as provided by the OpenNMT tool \citep{Klein2017}.
We apply the tool ``as-is" with its default translation hyper-parameters.
Note that our RNN system in this experiment is also an NMT-style model with an attention mechanism.

\noindent
\textbf{Scheduled Sampling Comparisons}:
\begin{table}
\centering
\begin{tabular}{ c c c }
\hline
Model & Dev & Test \\
\hline
{\SimpleRNN} & 76.85 \small{$\pm$0.39} & 73.52 \small{$\pm$0.36}\\
{\SCHRNN} & 76.93 \small{$\pm$0.32} & 73.65 \small{$\pm$0.29} \\
{\SCRNN} & 76.71 \small{$\pm$0.27} & 73.50 \small{$\pm$0.42} \\
{\OurRNN} & \textbf{77.10} \small{$\pm$0.29} & \textbf{73.82} \small{$\pm$0.29} \\
\hline
\end{tabular}
\caption{The adjusted actor-critic training compared to Scheduled Sampling (SS-RNN), and Self-Critical training (SC-RNN) on German NER.}
\label{scheduled-selfcritical-comparisons}
\end{table}
In the next experiment, we compare the adjusted actor-critic objective to Scheduled Sampling. We implement this approach using the inverse sigmoid schedule of \citet{scheduled} for annealing $\epsilon$, and denote it as SS-RNN. Table \ref{scheduled-selfcritical-comparisons} shows the comparison of SS-RNN with AC-RNN on German NER. Both systems improve over RNN, but AC-RNN is significantly better than SS-RNN with the P value of 0.040.
This result supports our hypothesis that the reinforcement-learning solutions should outperform Scheduled Sampling, as the adjusted actor-critic training considers the entire sequence,
whereas Scheduled Sampling addresses only exposure to the immediately previous token. We also observe that Scheduled Sampling, unlike the adjusted actor-critic training, is highly sensitive to the choice of sampling schedule (see the appendix for a quantitative analysis).

\noindent
\textbf{Self-Critical Comparisons}:
In our final experiment, we compare the adjusted actor-critic training with the Self-Critical policy training of \citet{Rennie2017SelfCriticalST} which does not require a critic model.
This method is intended to represent the state-of-the-art in reinforcement learning for sequence-to-sequence models with sequence-level rewards,
to be contrasted against our AC-RNN and its position-level rewards.
%\begin{table}
%\centering
%\begin{tabular}{ c c c c c }
%\hline
%Model & EnCh & EnJa & EnPe & EnTh \\
%\hline
%{\SimpleRNN} & 70.6  & 51.6 & 76.3 & 39.7\\
%{\SCRNN} & 70.2  & 51.8 & 77.2 & 41.3\\
%{\OurRNN} & \textbf{72.3} & \textbf{52.4} & \textbf{77.8} & \textbf{41.4} \\
%\hline
%\end{tabular}
%\caption[The Self-Critical training on transliteration]{The adjusted actor-critic objective compared to the Self-Critical policy training on the development sets of the NEWS-2018 shared task.}
%\label{dev-sc-vs-ac}
%\end{table}
%\begin{table}
%\centering
%\begin{tabular}{ c c c }
%\hline
%Model & Dev & Test \\
%\hline
%{\SimpleRNN} & 76.85 \small{$\pm$0.39} & 73.52 \small{$\pm$0.36}\\
%{\SCRNN} & 76.71 \small{$\pm$0.27} & 73.50 \small{$\pm$0.42} \\
%{\OurRNN} & \textbf{77.10} \small{$\pm$0.29} & \textbf{73.82} \small{$\pm$0.29} \\
%\hline
%\end{tabular}
%\caption[The Self-Critical training on German NER]{The adjusted actor-critic training compared to the Self-Critical policy training on German NER.}
%\label{de-ner-sc-vs-ac}
%\end{table}
The transliteration results shown in Table \ref{dev-trans} 
indicate that the Self-Critical training improves over RNN on EnJa and EnPe, and EnTh, 
however, it fails to beat AC-RNN across all the evaluation sets. 
On German NER, the Self-Critical training cannot improve over RNN as shown in 
Table \ref{scheduled-selfcritical-comparisons}. 
This observation is aligned with our initial hypothesis 
that the reinforcement-learning techniques applied to sequence labeling would benefit more 
from modelling the intermediate rewards, as is done with the Temporal Difference credits 
in the adjusted actor-critic training.
\section{Conclusion}
\label{conclusion}

We have proposed an adjusted actor-critic algorithm to train encoder-decoder RNNs for sequence labeling tasks.
Though related reinforcement-learning algorithms have previously been applied to sequence-to-sequence tasks,
our proposed AC-RNN 
is specialized to sequence-labeling 
by taking advantage of the per-position rewards. 
To our knowledge, we have presented the first direct, controlled comparison between CRFs and any form of RNN.
On NER and CCG supertagging, 
our system significantly outperforms both {\SimpleRNN} and {\crfRNN}, 
establishing the {\OurRNN} as 
an efficient alternative for sequence labeling.
We have also demonstrated the advantages of the AC-RNN in terms of its flexibility, fast training, and small memory footprint.
Finally, we showed that our proposal for handling exposure-bias outperforms
the related alternatives of Scheduled Sampling and Self-Critical policy training.

\vfill
\pagebreak

\bibliography{ref}
\bibliographystyle{iclr2019_conference}

\vfill
\pagebreak

\appendix
\label{append}

\section{Hyper-parameters}
Table \ref{hyper-detail} provides the hyper-parameters used in our experiments.

\begin{table}
\center
\small
\begin{tabular}{ |c|c|c|c| }
 \hline
 Hyper-parameter & En/De & CCG & TL\\
 \hline
 char\_embedding\_size & 32 & 32 & 128\\
 output\_embedding\_size & 32 & 128 & 128\\
 max\_gradient\_norm & 5.0 & 5.0 & 10.0\\
 encoder units & 256 & 512 & 256\\
 decoder units & 256 & 512 & 256\\
 batch size & 32 & 10 & 64\\
 $n$ & 2/4 & 8 & 6\\
 dropout& 0.5 & 0.5 & 0.5\\
 RNN gate & LSTM & LSTM & LSTM\\
\hline
\end{tabular}
\caption[The hyper-parameters used in the experiments]{The hyper-parameters used in the experiments.}
\label{hyper-detail}
\end{table}

\section{Schedule for Scheduled Sampling}
\begin{figure}
 \centering
  \includegraphics[width=10.00cm,height=6.96cm]{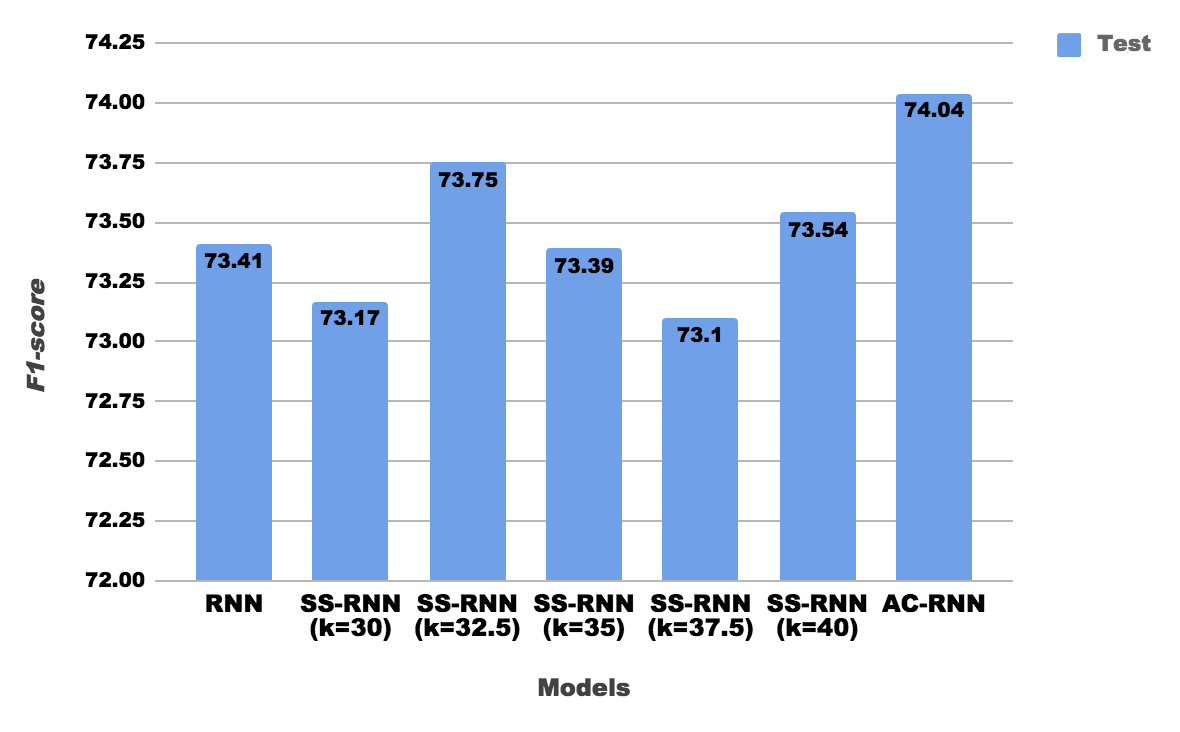}
  \vspace*{-.3in} 
 \caption[The effect of schedule on Scheduled Sampling]{The sensitivity of Scheduled Sampling on the choice of sampling schedule on German NER. Higher $k$ results in less sampling.}
 \label{ss-detailed}
 \end{figure}

Figure \ref{ss-detailed} illustrates that Scheduled Sampling is highly sensitive to the choice of sampling schedule.

\end{document}